# ADL4D: Towards A Contextually Rich Dataset for 4D Activities of Daily Living

Marsil Zakour*   Partha Pratim Nath*   Ludwig Lohmer
Emre Faik Gökçe   Martin Piccolrovazzi   Constantin Patsch
Yuankai Wu   Rahul Chaudhari   Eckehard Steinbach

Technical University of Munich, School of Computation, Information and Technology,
Department of Computer Engineering, Chair of Media Technology,
Munich Institute of Robotics and Machine Intelligence (MIRMI), Munich, Germany

`firstname.lastname@tum.de`

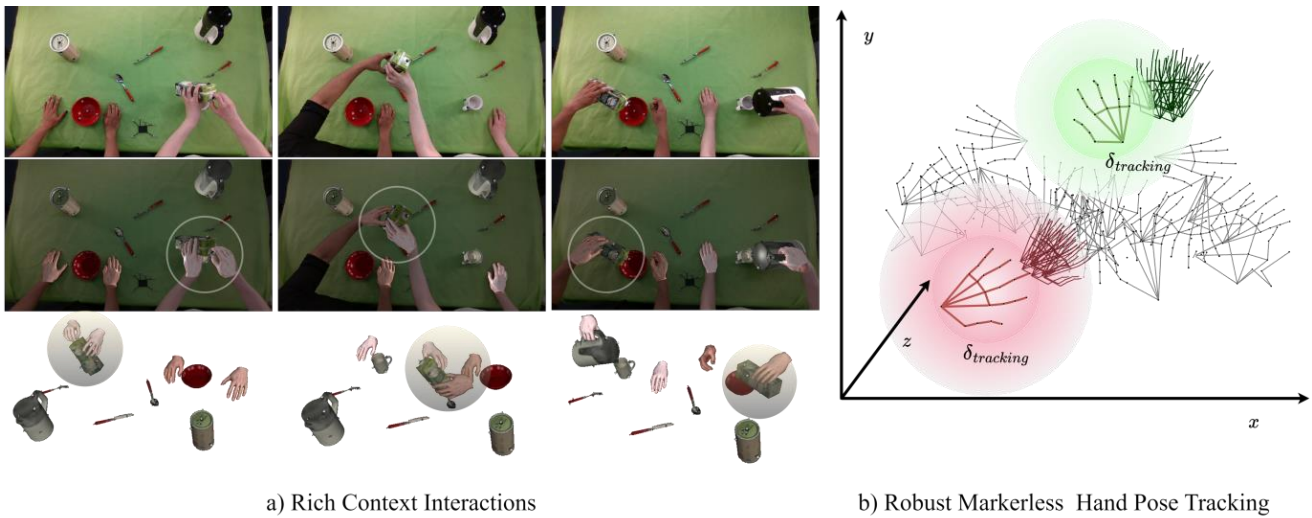

a) Rich Context Interactions

b) Robust Markerless Hand Pose Tracking

Figure 1. **Dataset and Annotation Setup Overview.** Figure a. Rich Context Interactions: An overview of our dataset Activities of Daily Living 4D (ADL4D). The top row shows an activity sequence view of our multi-view system. The middle row shows a rendering overlay of the object and hand poses. The bottom row shows a different point of view for the 3D scene. The subjects perform activities that require multiple objects while sharing some objects like the milk carton (highlighted with a circle in the middle and bottom rows). Figure b. Robust marker-less Hand Pose Tracking: The proposed region-growing approach for finding an inlier set of 3D hand proposals. The search is parametrized by the threshold $\delta_{tracking}$, which estimates the minimum hand displacement over consecutive frames.

## Abstract

*Hand-Object Interactions (HOIs) are conditioned on spatial and temporal contexts like surrounding objects, previous actions, and future intents (for example, grasping and handover actions vary greatly based on objects' proximity and trajectory obstruction). However, existing datasets for 4D HOI (3D HOI over time) are limited to one subject interacting with one object only. This restricts the generalization of learning-based HOI methods trained on those datasets. We introduce ADL4D, a dataset of up to two subjects interacting with different sets of objects performing Activities of Daily Living (ADL) like breakfast or lunch preparation activities. The transition between multiple objects to complete a certain task over time introduces a unique context lacking in existing datasets. Our dataset consists of 75 sequences with a total of 1.1M RGB-D frames, hand and object poses, and per-hand fine-grained action annotations. We develop an automatic system for multi-view multi-hand 3D pose annotation capable of tracking hand poses over time. We integrate and test it against publicly available datasets. Finally, we evaluate our dataset on the tasks of Hand Mesh Recovery (HMR) and Hand Action Segmentation (HAS).*

## 1. Introduction

Activities of Daily Living (ADL) involve humans interacting with different objects in their surroundings, transitioning from one set of objects to the next in order to execute a structured plan. Hands and object poses convey most of the information about activities. For example, the knowledge about the hand and a milk bottle pose can differentiate between the "pour milk" and "open milk" actions. Therefore, a 4D understanding of these interactions requires data with hands, object poses, and actions over time. Recently, several new Hand-Object Interaction (HOI) datasets were introduced. However, these datasets do not cover full activities, rather only single or few consecutive interactions like pickup box, open milk, handover, read book, etc. [2, 33, 57], or they cover consecutive interactions including a single object with no long-term plan or action annotations [13]. Dexterous hand-object interaction sequences are key for tasks like robot learning from demonstration; for example, Chao et al. [2] used hand-object interaction data to learn to predict plausible clamp grasp from hand-object images. Another important application is the generative modeling of hand-object animations. For example, He et al. [59] generate hand interaction poses given hand-object distance information and trajectories. Similarly, Taheri et al.[51, 52] use hand object datasets for grasp and pickup action generation. These use cases are currently limited by existing datasets complexity and cannot be generalized to more complex scenarios. Given the existing dataset limitations, we introduce our dataset ADL4D which we curate towards contextually rich activities, and at-length interactions of multiple objects and subjects. A sample of ADL4D is shown in Figure 1 (a). The need for further complex data with more hands and sophisticated object interactions promotes the need for a more robust tracking system. The challenge with activity sequences is maintaining consistent 3D tracking of hands over time. Existing marker-free automatic approaches use a high frame rate of 60 to 90 FPS during annotation for a short duration and assume a maximum of two hands only [20, 21]. Alternatively, using a very large number of cameras [31, 32, 41, 49] would increase the robustness of the systems, the scale comes with a high cost, synchronization, and hardware requirements. This will limit the number of interactions that could occur in one sequence. Unlike multi-person pose estimation and tracking problems, hands are visually more similar and have finer articulation which results in less continuity in trajectories.

In our method, we build on top of the Dynamic Matching (DM) component from Huang et al. [25], to robustly estimate 3D hands in sequences. We have new data patches continuously added to the system, therefore, we develop a model-ensemble method. This facilitates fine-tuning the 2D detection models on the new data patches only.

We run exhaustive experiments on the part of our data that is annotated through human supervision, as well as other publicly available datasets [2] and [33].

Our contribution consists of 3 main parts

- A dataset with 1.1M images (75 long sequences for breakfast and lunch activities). These sequences include up to 2 subjects interacting with multiple objects. The annotations include hand, object, 3D poses, and hand-level fine-grained annotations.
- An automatic annotation tool (with semi-automatic GUI support) compatible with different existing datasets.
- An extensive analysis of the automatic hand tracking method across different datasets and an evaluation of our proposed dataset (ADL4D) across tasks like Hand Action Segmentation (HAS) and Hand Mesh Recovery (HMR).

## 2. Related Work

### 2.1. Datasets for 3D Hand-Object Activities

There have been many datasets capturing 3D human interactions with surrounding objects. These datasets vary in the richness of annotations, data source, and annotation method. Table 1 shows an overview of recent datasets covering one or both areas. Similar datasets such as those referenced in [20], and [48] are not included in the table as they are superseded by a newer version. The existing work could be classified into two major types. First, datasets with a focus on long and complex tasks with multiple objects and a single subject like assembly tasks in [44, 46], as shown in Table 1. These dataset provide 2D or 3D hand annotations, but no 3D object annotations. Second, HOI datasets like [2, 13, 21, 33, 37] are mainly focusing on short interactions with one object. Table 1 shows that these datasets have action annotations however, there is no transition between different objects in order to accomplish a higher level task.

Due to self-occlusions and inter-instance occlusions, accurate acquisition of hand-object poses is a very challenging problem. Mono and stereo camera systems in datasets like [17, 37] would require human annotation or additional sensor data. Therefore, multiple datasets follow a multi-view system [2, 21, 33, 44]. These annotation methods extract semantic information from 2D views and fit hand and object parametric models to them [3, 47]. These systems are limited by the number of objects actively moving in the scene as they are prone to failure. For example, [2] is limited to one hand and one moving object with multiple static objects in the scene. Alternatively, [21, 33] are limited to 2 hands and one object only. A more reliable approach is to use marker-based motion capture systems, for example, Yang et al.[57] use Optitrack cameras to acquire the object pose and annotate hand 2D key points through crowd-sourcing services. Similarly, Fan et al. [13] use Vicon Mocap cameras to fully annotate hands and objects. The main issue with the latter approach is that, unlike objects, hand keypoint estimation methods should generalize to unseen hands, and the existence of markers will bias the model to detect the markers instead. Therefore, there have been attempts to mitigate this issue by erasing the markers using generative image models [56]. Based on this information, we annotate the objects using a marker-based system and implement our own method to semi-automatically annotate the hands with no markers.

## 2.2. Hand Mesh Recovery Methods

There have been many publications into the work of HMR from single RGB images. Existing methods can be broadly divided into two approaches, model based and model free methods. Jiang et al. [29] attempts to combine the two by using a MANO[47] model as a prior net to capture the probability distribution of hand joints and mesh vertices. Their Attention based Mesh Vertices Uncertainty Regression(AMVUR) model learns the conditioned probability distribution of joints and vertices based on the MANO priors, allowing representation of hand poses larger than the MANO subspace.

**Model Based Methods:** These approaches attempt to reconstruct the 3D hand by regressing the pose and shape parameters of a parametric hand model like MANO [47]. MANO defines a hand mesh captured by a differentiable parameter space of 61 values learned from over 1000 high-resolution 3D hand scans. Zhang et al. [60] utilizes an iterative regression approach to fit model parameters from 2D joint heatmaps.

Transformers are a good approach in robustly estimating hand features from a Convolutional Neural Network (CNN) like Resnet [23] embeddings. Park et al. [45] regress a MANO model onto these features to generate the hand mesh. Hasson et al. [22] provide an end-to-end learnable model that predicts a MANO hand model and a reconstructed interacting object from Resnet18 [23] embeddings, and regresses the poses together using a novel contact loss for realistic hand mesh recovery.

**Model Free Methods:** Hampali et al. [21, 35] utilises transformer network on 2D heatmaps and embeddings from by a CNN to generate a 3D handpose. Moon et al. [40] creates an image-to-lixel prediction network using a PoseNet[6] to predict 3-dimensional lixel based heatmaps of every joint. This is fed as input to a Meshnet model that predicts the 3-dimensional lixel heatmaps of every vertex of the hand mesh. Graph Convolutional Network (GCN) are also demonstrated as a reliable approach to estimate hand meshes directly. For example Choi et al. [9] utilises a PoseNet [6] to estimate relative 3D joints and a GCN to estimate the 3D human and hand mesh vertices. Ge et al.

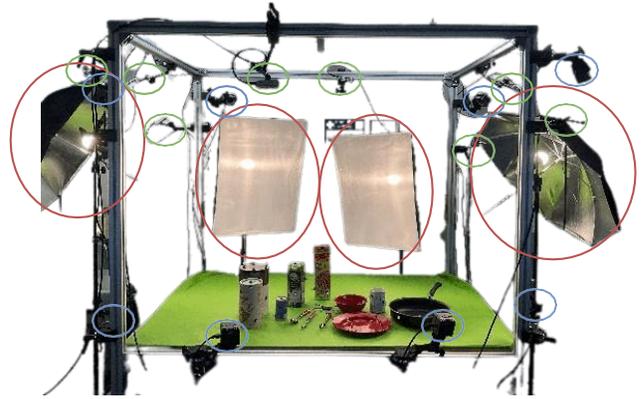

Figure 2. **Tabletop Setup.** Our setup consists of 8 RealSense D435 cameras (green ellipses), 8 Optitrack Prime 13 X cameras (blue ellipses), and 4 spotlights (red ellipses). On the tabletop, we show the interaction objects

[18] uses a GCN to directly regress the mesh vertices onto heatmaps generated by stacked hourglass network [42].

Also works in 3D hand pose estimation like [28, 61], can easily be extended to generate hand meshes using parametric model regression.

## 2.3. Activity Recognition and Segmentation Methods

Action recognition is an important part of human-centered computer vision. Short-term action classification model average 3D temporal convolutions [5][16][16]. However, these models have a limited clip duration that they can process. Models segmenting longer sequences of videos leverage the latter classification models as feature extractors [34][14][58]. Alternatively, action detection based on hand 2D and 3D gained additional importance due to its applications in AR/VR [43] [33] [48].

## 3. Data Acquisition and Annotation Setup

### 3.1. Data Recording

Figure 2 shows our data recording setup. We use 8 RGB-D RealSense cameras and 8 Optitrack cameras for object tracking. Since our activity sequences could last up to 2 minutes, reducing blurry frames is essential for smooth hand tracking. Even though we are using D435 model of cameras that advertise a global shutter. However this does not apply to the color imager on the D435 which is a rolling shutter. In our recordings, we notice that normal ambient daylight and indoor light conditions produce multiple blurry images. In order to solve this issue while maintaining a stable frame rate, we use 4 spotlights with diffusers and con-

| Dataset | Modality | Funct. intent | Action. annot | 3D Hands | 6D Object | #Obj. per seq | Hands per seq | Mul. Subj. Inter | #Sub. | #Obj. | #Env. | #Seq. | #Img | #Cam | Resolution |
|---|---|---|---|---|---|---|---|---|---|---|---|---|---|---|---|
| FPHA [17] | RGB-D | Y | Y | Y | Y | 1 | 1 | N | 6 | 4 | 3 | 1.2K | 105K | 1 | 1920x1080 |
| DexYCB [2] | RGB-D | N | N | Y | Y | 1 | 1 | N | 10 | 20 | 1 | 1K | 582K | 8 | 640x480 |
| H2O [33] | RGB-D | Y | Y | Y | Y | 1 | 2 | N | 4 | 8 | 3 | 24 | 571K | 5 | 1280x720 |
| H2O3D [21] | RGB-D | N | N | Y | Y | 1 | 2 | N | 6 | 10 | 1 | 65 | 76K | 5 | 640x480 |
| MECCANO [46] | RGB-D | Y | Y | N | N | 20 | 2 | N | 20 | 20** | 2 | 20 | 300K | 1 | 1920x1080 |
| OakInk [57] | RGB-D | Y | Y | Y | Y | 1 | 2 | Y | 12 | 32 | 1 | 1.3K | 230K | 4 | 640x480 |
| HOI4D [37] | RGB-D | Y | Y | Y | Y | 1 | 1 | N | 9 | 20 | 610 | 5K | 2.4M | 1 | 1280x800 |
| AssemblyHands[44] | RGB/BW | Y | Y | Y | N | - | 2 | N | 34 | - | 1 | 4.3K* | 3M | 8+4 | 1080p/480p |
| ARCTIC [13] | RGB | Y | N | Y | Y | 1 | 2 | N | 9 | 11 | 1 | 339 | 2.1M | 9 | 2880x2000 |
| ADL4D (ours) | RGB-D | Y | Y | Y | Y | 4-12 | 2-4 | Y | 8 | 12 | 1 | 75 | 1.1M | 8 | 1920x1080 |

Table 1. **Related Datasets Comparison** The action column denotes presence of action annotations if any. Number of objects Per seq. denotes range of objects present and used in a single recording. *Assembly101[48] with 4.3K sequences and 20M images are subsampled to generate Assembly Hands. Data is not provided if any sequence in whole is excluded. **Meccano sequnces consist of a single toy assembly from 19 unique components, and 2 tools. Which are altogether grouped into 20 classes

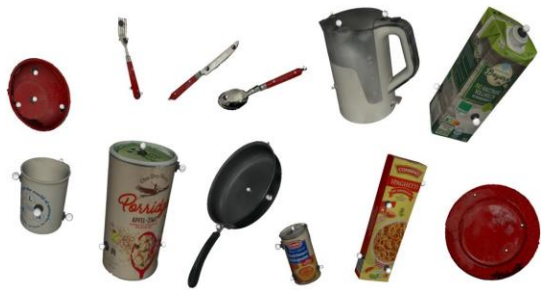

Figure 3. **Object Models.** The set consists of 12 high-quality textured meshes with their markers

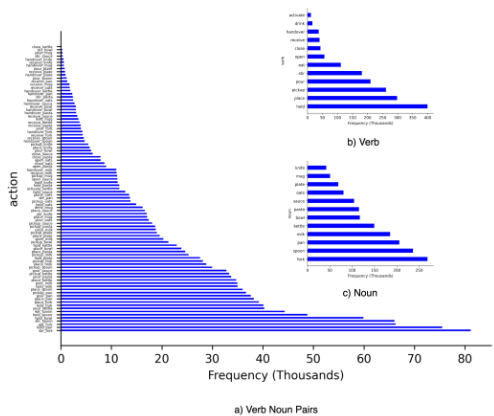

Figure 4. **Tail Action Distribution.** A histogram plot of frame-wise action distributions sorted by occurrence frequency. Figure(a). Shows Verb-Noun pairs co-occurrence frequency. Figure (b) shows verb frequencies. Figure (c) shows Nouns (Interaction Objects) frequencies.

figure the RealSense cameras with a fixed exposure time to the smallest possible value while no frames are blurry (e.g. 0.2 ms). We collect the RGB-D data at 30 FPS and the Optitrack cameras at 120 FPS.

### 3.2. Object Meshes

The used models are shown in Figure 3. We use a mix of YCB [4] and our own objects in order to have meaningful activity scenarios. The objects are scanned using two high-resolution structured-light scanners [1] that are suitable for small and medium object sizes. We provide the textured meshes with 3 different resolutions. We ensure that the YCB [3] scans are aligned with the original models.

### 3.3. Action annotations

We use VIA software [12] to annotate videos. Since the videos are software-synchronized, we annotate one view and apply the labels to other views. In 4, we plot the frequency of verbs, nouns, and actions (verb-noun pairs) in multiples of 1 thousand. In sub-figure (b), we observe that verbs that initiate interactions (pickup) and finish interaction (place) are the most common among verbs. Meanwhile, in sub-figure (a), we can notice that actions with repeating patterns (stir and eat) occupy many frames. Similarly, in sub-figure (c), multi-functional objects like spoon and fork, which correspond to (eat, stir) verbs, are the most frequent among nouns.

### 3.4. Calibration and Synchronization

We use a checkerboard pattern to calibrate the RealSense cameras and ensure the highest re-projection error is less than 0.05 pixels at 1280x720 resolution. By annotating markers manually, we align the transformation trees of the RGB-D cameras to the mocap system world coordinate frame to form one transformation tree. We also align the mesh scans to their markers by picking 3D points on the spherical markers' surface and clustering them such that each cluster centroid matches the marker centroid. The different recorded sequences are then synchronized and sub-sampled at 20 FPS. We find that at 20 FPS is a good balance

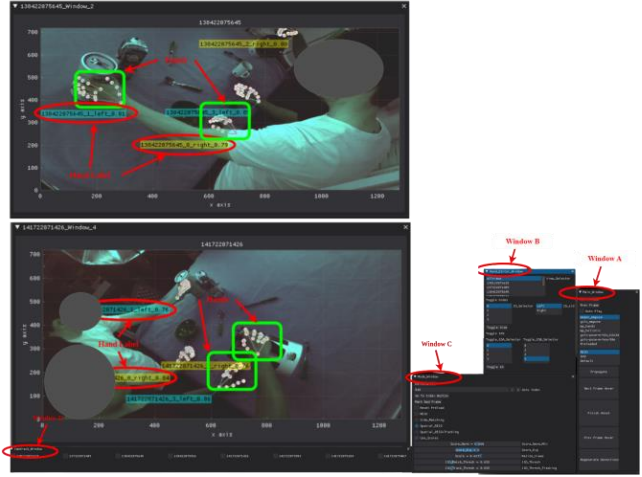

Figure 5. **Hand Pose Annotation GUI.** We also support semi-automatic annotation mode with live human supervision and control.

between redundancy and smooth image stream.

### 3.5. Hand Pose Annotation and Tracking

We take into consideration two aspects regarding the annotation method. Firstly, the annotated sequences are long, 1-2 minutes, synced at 20 fps. Secondly, We have 1-2 subjects in each sequence. Therefore, we developed our annotation method to support automatic and semi-automatic annotation. The early stages of data collection will require more careful annotation with human supervision since key point estimators are not curated toward the data distribution. Figure 5 shows the annotation GUI. The usage of the GUI is detailed in 3.6.1

In Figure 6 we visualize the PCA components of the hand poses of ADL4D, DexYCB[2], and H2O [33]. The visualization suggests that the poses in ADL4D are more diverse while sharing similarities with poses in DexYCB and H2O. We integrate multiple existing datasets into our system, including [2, 33]. We use these two datasets for evaluating our hand annotation. We train/use different keypoint estimators [30][55][38][11] on different dataset including [7] [33] [41], and incrementally add our own data patches. Since we may have multiple subjects in each scene, hand-side information is not enough to uplift 2D detections to 3D. Therefore, we extend over the idea of DM by [25] to isolate 3D hands. We pair this with a weighted averaging of the detected side classes to robustly estimate the side from all matched detections.

### 3.6. 3D Pose Subspace Clustering

The DM[25] algorithm is a clustering approach applied to the 3D pose space obtained by pairwise uplifting of every detection from all views. Given a threshold $\delta_{dynamic}$, it searches for all clusters of detection pairs whose 3D pose lies within $\delta_{dynamic}$ of each other. This algorithm has a 3-camera dependency, such that it requires a single hand to be detected in atleast 3 camera views to isolate it into an output cluster. While we observe this approach to work well with finetuned models on existing datasets, our initial batches of data annotation are often unable to isolate all hands of users, primarily due to insufficient detections caused by lower detector performance. Therefore, in order to robustly uplift 2D detections to 3D, we also extend the DM algorithm with temporal information from the previous frames to reduce the 3-camera dependency to a 2-camera dependency. We refer to this as the Tracking Mode (TM) in our paper. In DM, the association to the previous frame is done based on spatial proximity post-cluster estimation.

The DM and TM approaches search clusters within a certain radius in the 3D space. Due to varying speed of hand movement or factors affecting 2D detector accuracy, a fixed threshold may however fail to annotate all frames for a full sequence. Cases for these include when two hands approach closer together but move slowly; a smaller threshold is required to prevent incorrect clustering compared to when a hand starts moving or has a sudden change in velocity or trajectory during a sequence. We, therefore, design and compare two search approaches with different matching criteria (discussed in 3.6.2) for evaluating the quality of the clusters over a given specific threshold range for $\delta_{tracking}$ or $\delta_{dynamic}$.

#### 3.6.1 Semi-Auto Annotation-Manual Searching

As shown in Figure 5, we support a manual search method for our clustering parameters and overlay results on a GUI where the multi-camera images are shown concurrently, with overlaid hands detection. There are 4 primary control windows providing access to different parameters of our application. Window A: The Main Window provides access to incremental or automated frame steps forward or back, the model selection from detector zoo, overlay configuration between raw 2D detections, matched 2D detections or re-projected 3D estimation. Window B: Is provided in case a user wishes to manually match hand indices. Window C: Provides controls to the 3D matching parameters. Window D: Provides a camera filtering option to reject detections.

#### 3.6.2 Search Matching Criterion and Auto Annotation

The search method will grow the set of inliers by searching in the neighborhood of the accepted threshold $\delta_{dynamic}$ or $\delta_{tracking}$ found from the previous frame in the sequence. We incrementally increase the magnitude of the offset and pick a result based on the matching criteria. We evaluate three different matching criteria:

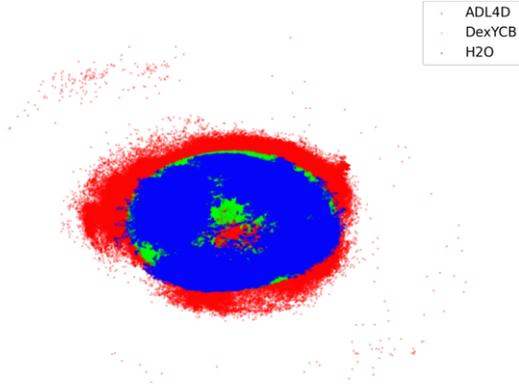

Figure 6. **Hand Pose Variation.** A PCA visualization of the first two principal components of the hand poses. The plot shows that our dataset ADL4D poses (red) are more diverse than DexYcb [2] H2O [33] (blue)

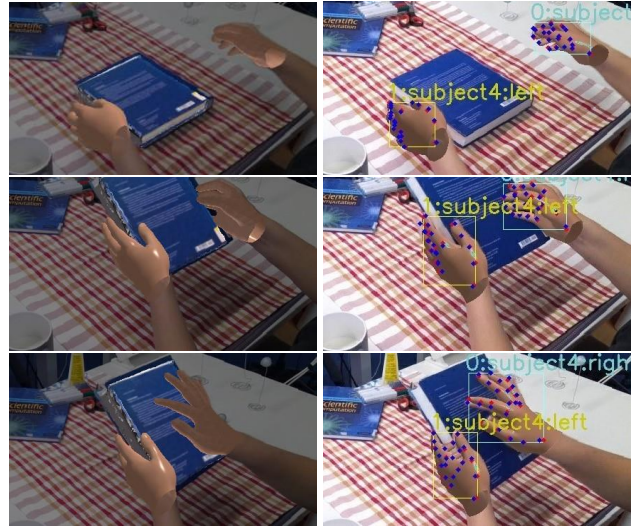

Figure 7. **H2O Annotation (Best shown with color).** A comparison between H2O [33] Ground Truth (left) and our sequence annotation (right). We render the corresponding Mano[47] Mesh to one of the views. The red keypoints are the hand-tracking output, while the blue keypoints correspond to optimised Mano joints. From a qualitative perspective there is almost no observable difference to H2O GT and our annotation.

- Re-projection Error (Repr), this approach filters clusters based on a cluster size threshold. We denote the set of accepted clusters as $C_{large}$. This approach will select the re-projection-error minimizer $c^*$ as described in Equation 1, where $\pi$ projects the 3D joints to each camera space.

$$c^* = \mathrm{argmin}_{c \in C_{large}} ||\pi(J^c_{3D}) - J^c_{2d}||^2_2 \qquad (1)$$

- Closest Displacement (CD) this searches from the last accepted threshold with increasing offsets until the first valid set of clusters for all hands is returned.
- No Search (NS) an average threshold with no search performed.

The search criteria algorithms are detailed in the supplementary material.

In the case where the search criteria fail to find a valid cluster, the last successful cluster is used instead.

## 4. Experiments

### 4.1. Annotation Method Experiments

We perform a detailed ablation study of the matching criteria introduced in section 3.6.2 to evaluate their impact on the annotation accuracy. We perform this on subsets of the test split of 3 different datasets in increasing order of complexity. DexYCB[2] H2O[33], and our dataset ADL4D. For each study a HRNet model[55] is trained on the train splits of the respective dataset. To evaluate 3D error, we report Mean Per-Joint Position Error (MPJPE) in millimeters, Percentage of Correct Keypoints - Area Under the Curve (PCK-AUC) of each approach. Also, we evaluate the tracking accuracy for 3D keypoint trajectories over time by reporting the keypoint tracking accuracy[39] at 1 cm, the number of total skipped(unannotated) frames. For completeness, we also report the 2D COCO Keypoint and bounding box evaluation metrics [36] after projecting the joints from 3D to 2D.

### 4.2. Ablation Study on ADL4D

We conduct an ablation on all possible combinations of the search methods introduced in section 3.6 and the search matching criteria introduced in 3.6.2 . This will enumerate six possible configurations. We automatically annotate the single-subject test sequences (4 sequences) in ADL4D using each combination. The results of the experiment can be found in Table 3. We observe a significant improvement across all metrics when comparing NS-DM to NS-TM, with a 50% reduction in skipped frames and a 12% increase in tracking accuracy. With the implementation of the Repr, we see a further reduction of 80% in skipped frames and a 10% increase in tracking accuracy.

### 4.3. Tracking Evaluation Across Multiple Datasets

In this experiment, we evaluate the performance of our annotation method on the sequences of DexYCB and H2O introduced at the beginning of the section 4.1. We evaluate the TM search method performance and report the results in Table 2. From Table 2 and the TM results in Table 3 we can observe that the search method has a marginal impact on simple and short sequences like the ones in DexYCB. For a slightly more difficult sequence like the ones in H2O, we can observe a slight drop in the number of skipped frames (un-annotated frames) when using a search method compared to NS. The impact of the search methods CD and Repr increases with more complex sequences like the ones in ADL4D reported in Table 3. We can also observe that the search criterion is inconsistent among different datasets. We hypothesize that this could be related to multiple factors, such as the number of hands, views, and frame rate. Therefore, we consider that the search criteria hyper-parameters like cluster size threshold should be tuned for every multi-camera system separately.

| Method | Criterion | MPJPE (abs) | AUC (abs) | $mAP_{Box}$ | | $mAP_{Kpt}$ | | Skipped Frames | Track Acc. @1cm |
|---|---|---|---|---|---|---|---|---|---|
| | | | | 50% | 50:90% | 50% | 50:90% | | |
| TM | NS | 5.18 | 0.8963 | 0.9900 | 0.7837 | 0.9759 | 0.7768 | 19 | 0.8353 |
| H2O | **CD** | **5.15** | **0.8968** | **0.9899** | **0.7846** | **0.9759** | **0.7783** | **0** | **0.8367** |
| | Repr | 5.79 | 0.8842 | 0.9845 | 0.7560 | 0.9565 | 0.7317 | 3 | 0.7634 |
| TM | NS | 6.64 | 0.8671 | 0.9949 | 0.7690 | 0.8022 | 0.4414 | 0 | 0.6463 |
| DYCB | CD | 6.64 | 0.8671 | 0.9949 | 0.7690 | 0.8022 | 0.4414 | 0 | 0.6463 |
| | **Repr** | **6.62** | **0.8676** | **0.9899** | **0.7699** | **0.7944** | **0.4518** | **0** | **0.6405** |

Table 2. **3D Annotation Experiment DexYCB++H2O(10 seqs each).** H2O contains 6654 total frames and DexYCB is 720 frames total

| Method | Criterion | MPJPE (abs) | AUC (abs) | $mAP_{Box}$ | | $mAP_{Kpt}$ | | Skipped Frames | Track Acc. @1cm |
|---|---|---|---|---|---|---|---|---|---|
| | | | | 50% | 50:90% | 50% | 50:90% | | |
| DM | NS | 18.30 | 0.7952 | 0.7027 | 0.6121 | 0.6861 | 0.6123 | 2005 | 0.6261 |
| | CD | 18.30 | 0.7952 | 0.7027 | 0.6121 | 0.6861 | 0.6123 | 2005 | 0.6261 |
| | Repr | 18.33 | 0.7947 | 0.6978 | 0.6101 | 0.6861 | 0.6112 | 2005 | 0.6261 |
| TM | NS | 13.60 | 0.8326 | 0.8410 | 0.7312 | 0.8172 | 0.7279 | 1049 | 0.7413 |
| | CD | 8.47 | 0.8769 | 0.9319 | 0.8043 | 0.9017 | 0.7974 | 332 | 0.7971 |
| | **Repr** | **5.94** | **0.8952** | **0.9596** | **0.8323** | **0.9371** | **0.8328** | **226** | **0.8433** |

Table 3. **3D Annotation Experiment 4 ADL4D sequences** with 6815 total frames. Our region growing approach combined with tracking mode demonstrates a significant improvement in percentage of annotated frames over the Dynamic Matching, resulting in better absolute MPJPE and Tracking score

Finally, we visualize qualitative results on our annotation method with TM - CD on one sequence of H2O 7.

### 4.4. HMR from a Monocular RGB Image

We run a cross-dataset evaluation against [2] and [33] where we use one dataset for training and testing on others. We use the standard protocol for HMR evaluation introduced by [10], and compare the performance using root-relative MPJPE. In Table 4 we show the results of the Lixel model [40] (hand-model free) while in Table 5 show HandOccNet [45] (parametric model). We can see from Table 4 that models trained on ADL4D and tested on H2O are always better than the ones trained on DexYCB. The min cross dataset error is 33.34 mm (in Table 4) when testing ADL4D on H2O. We also can observe that the results on DexYCB are bad for both training and testing on DexYCB in Table 4 and Table 5. This suggests that the data distribution is far away from ADL4D and H2O. We suspect that this could be due to the blurry, low-resolution images in DexYCB. In Figure 8, we show qualitative results from the model trained on ADL4D to other datasets.

| Train/Test | DexYCB[2] | H2O [33] | ADL4D |
|---|---|---|---|
| DexYCB[2] | 12.48 | 44.96 | 48.94 |
| H2O [33] | 54.76 | 14.35 | 40.89 |
| ADL4D | 54.67 | 33.34 | 13.18 |

Table 4. **I2L Meshnet[40] Cross-Dataset Evaluation.** Cross dataset evaluation against DexYCB and H2O datasets the reported metrics are root relative MPJPE in millimeters

| Train/Test | DexYCB[2] | H2O [33] | ADL4D |
|---|---|---|---|
| DexYCB[2] | 13.88 | 39.27 | 47.51 |
| H2O [33] | 54.58 | 13.28 | 35.36 |
| ADL4D | 54.73 | 35.16 | 13.19 |

Table 5. **HandOccnet[45] Cross-Dataset Evaluation.** Cross dataset evaluation against DexYCB and H2O datasets the reported metrics are root relative MPJPE in millimeters

### 4.5. Hand Action Segmentation

We evaluate the performance of hand pose information for action segmentation. The hand pose could be used as a common input modality for deep learning models independent of the input sensor (color-image, depth-image, or wearable glove). In this experiment, we use the verb labels for prediction as we are interested in the hand pose as input. When evaluating Action Segmentation (AS) models on datasets like [15, 37, 50], the TCN family of models have shown good performance, and different variations of these models like the one referenced in [34, 58] are among the state of the art models. We use an ASFormer to aggregate features, generate frame-wise verb predictions, and evaluate the segmentation frame-wise accuracy, edit distance, and F1 score with overlap thresholds 10%, 25%, and 50%. We extract I3D [5] using RGB and Flow. To support hand-level predictions, we generate a video of each hand trajectory using the hand bounding box and the hand image. Figure 9 illustrates the feature extraction pipeline. The image is padded to have equal width and height. We use the video-features library [26] to compute the I3D features based on Raft flow [53] and RGB streams. We use I3D features as a baseline and compare it to the hand pose predicted by an HMR model [45]. We report the performance difference in Table 6. We can observe that ASFormer performance is better when using pose features for both frame-wise accuracy and segment-wise metrics. We suspect that the better performance can be attributed to two reasons. First, the hand pose values are local joint rotations, which will have large changes in values only during action transitions and dexterous manipulation. This "locality" will help the TCN to better detect events. Secondly, the dynamic background in the video generated by the hand trajectory is unsuitable and would generate noisy I3D features due to the constant motion and background changes. Finally, we note that the frame-wise accuracy and F1 score values suggest that our dataset ADL4D is challenging for the HAS task. We hypothesize that this is due to the fine-grained actions with long sequences and complex temporal contexts.

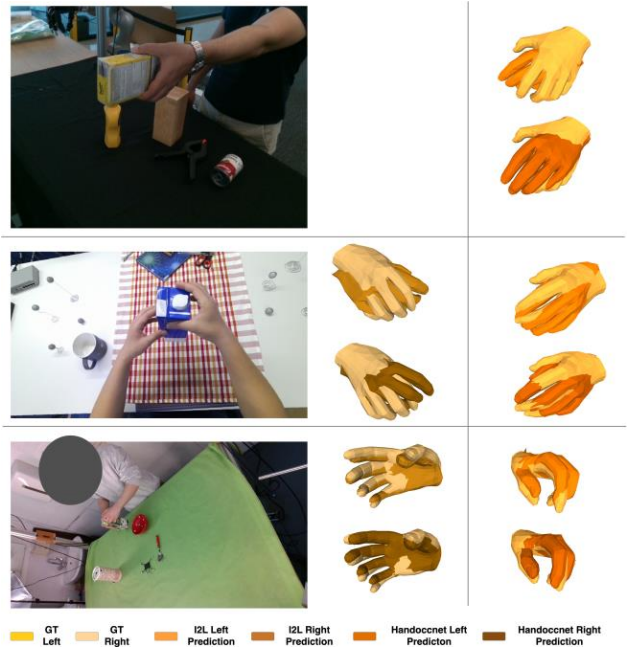

Figure 8. **Cross Dataset HMR Qualitative Results.** The input images, from DexYCB [2], H2O[33], and ADL4D in order. Odd rows correspond to I2L [40] and Even rows correspond to HandOccNet [45] both trained on ADL4D.

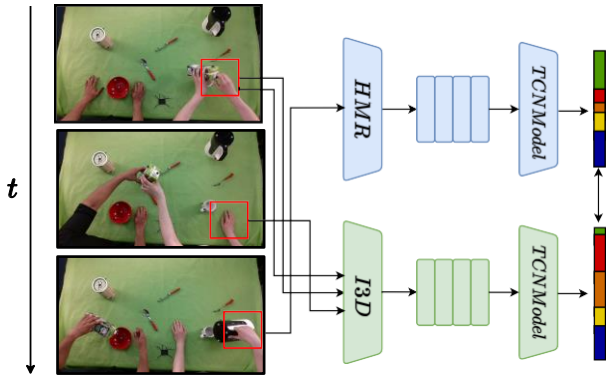

Figure 9. **Hand Action Segmentation Pipeline.** An illustrative figure for our HAS feature extraction pipeline. We use the hand-bounding box trajectory as input to apply hand-level action segmentation. As a baseline, I3D features are extracted based on RGB and Flow. We compare this against hand pose predictions from [45]. ASFormer is used as a Temporal Convolution Network (TCN).

| Features | Acc. | Edit | F1@10,25,50 | | |
|---|---|---|---|---|---|
| I3D | 32.77 | 41.66 | 24.59 | 18.21 | 7.12 |
| Pose [45] | **55.04** | **52.66** | **55.01** | **33.78** | **33.78** |

Table 6. **Hand Action Segmentation using ASFormer [58]**

## 5. Conclusion

We introduce ADL4D, a dataset for 4D human activities. Unlike existing HOI datasets, we focus on rich context long-term activities with multi-subject and multi-object interactions. We evaluated the dataset in the areas of HMR and HAS. We also demonstrated the effectiveness of our marker-free hand annotation system for long sequences and crowded scenes.

### 5.1. Limitations and Future Work

As we use an aggregate over multiple 2D keypoint estimators in our method, hence we use SVD to triangulate the keypoints. The results can still be improved by using a learnable triangulation approach [27, 44], or enhancing the 2D keypoints using Epipolar Transformers [24]. However, we argue that in the early stages of data collection, the deep-learning-based methods have a bad cross-dataset generalization [44], and they have to be trained on similar data distribution. We think that our annotation pipeline can help with accelerating the annotation process for similar datasets, especially with the support of our semi-automatic, and automatic annotation setups. Regarding our proposed dataset, we think that ADL4D will facilitate a new range of applications that was not possible before. For example, using hand-level action annotations which are missing in the existing dataset, ADL4D could help transition action to motion models [8, 19, 54] where existing generative action-based HOI animations are limited to pick up, and grasp, or handover actions [52][57].